\documentclass[conference]{IEEEtran}
\IEEEoverridecommandlockouts
\usepackage[utf8]{inputenc} 
\usepackage[T1]{fontenc}    
\usepackage{hyperref}       
\usepackage{url}            
\usepackage{booktabs}       
\usepackage{amsfonts}       
\usepackage{nicefrac}       
\usepackage{microtype}      
\usepackage{xcolor}         
\usepackage{graphicx}
\usepackage{amsmath,amssymb}
\usepackage{rotating}
\usepackage{colortbl}       
\usepackage{multirow}       
\usepackage{array}          
\usepackage{makecell}       
\usepackage{siunitx}        
\usepackage{placeins}       
\usepackage{float}          
\usepackage{tabularx}
\usepackage{caption}

\def\BibTeX{{\rm B\kern-.05em{\sc i\kern-.025em b}\kern-.08em
    T\kern-.1667em\lower.7ex\hbox{E}\kern-.125emX}}
\begin{document} 

\title{Transforming Expert Knowledge into Scalable Ontology via Large Language Models}

\author{\IEEEauthorblockN{Ikkei Itoku}
\IEEEauthorblockA{\textit{Amazon}\\
New York, USA \\
itoku@amazon.com}
\and
\IEEEauthorblockN{David Theil}
\IEEEauthorblockA{\textit{Amazon}\\
Arlington, USA \\
dtheil@amazon.com}
\and
\IEEEauthorblockN{Evelyn Eichelsdoerfer Uehara}
\IEEEauthorblockA{\textit{Amazon}\\
Seattle, USA \\
eeichels@amazon.com}
\and
\IEEEauthorblockN{Sreyoshi Bhaduri}
\IEEEauthorblockA{\textit{Amazon}\\
New York, USA \\
drsre@amazon.com}
\and
\IEEEauthorblockN{Junnosuke Kuroda}
\IEEEauthorblockA{\textit{Amazon}\\
Seattle, USA \\
kurodaju@amazon.com}
\and
\IEEEauthorblockN{Toshi Yumoto}
\IEEEauthorblockA{\textit{Amazon}\\
Arlington, USA \\
futosy@amazon.com}
\and
\IEEEauthorblockN{Alex Gil}
\IEEEauthorblockA{\textit{Amazon}\\
New York, USA \\
gilalexg@amazon.com}
\and
\IEEEauthorblockN{Natalie Perez}
\IEEEauthorblockA{\textit{Amazon}\\
Honolulu, USA \\
natkper@amazon.com}
\and
\IEEEauthorblockN{Rajesh Cherukuri}
\IEEEauthorblockA{\textit{Amazon}\\
Seattle, USA \\
rccheruk@amazon.com}
\and
\IEEEauthorblockN{Naumaan Nayyar}
\IEEEauthorblockA{\textit{Amazon}\\
Seattle, USA \\
nayyarnn@amazon.com}
}

\maketitle

\begin{abstract}
%
Having a unified, coherent taxonomy is essential for effective knowledge representation in domain-specific applications as diverse terminologies need to be mapped to underlying concepts. Traditional manual approaches to taxonomy alignment rely on expert review of concept pairs, but this becomes prohibitively expensive and time-consuming at scale, while subjective interpretations often lead to expert disagreements. Existing automated methods for taxonomy alignment have shown promise but face limitations in handling nuanced semantic relationships and maintaining consistency across different domains. These approaches often struggle with context-dependent concept mappings and lack transparent reasoning processes. We propose a novel framework that combines large language models (LLMs) with expert calibration and iterative prompt optimization to automate taxonomy alignment. Our method integrates expert-labeled examples, multi-stage prompt engineering, and human validation to guide LLMs in generating both taxonomy linkages and supporting rationales. In evaluating our framework on a domain-specific mapping task of concept essentiality, we achieved an F1-score of 0.97, substantially exceeding the human benchmark of 0.68. These results demonstrate the effectiveness of our approach in scaling taxonomy alignment while maintaining high-quality mappings and preserving expert oversight for ambiguous cases.
\end{abstract}

\section{Introduction}
In today's data-driven world, organizations evolve and expand their operations across multiple domains and functions. Knowledge management is essential for organizations to scale efficiently, and ontologies and taxonomies are important tools within a knowledge management system (\cite{Osman2022}. Ontologies and taxonomies are often used to extract, organize, and structure knowledge across domains and fields such as healthcare, business or education to allow consistent classification, retrieval, and information integration (\cite{abu2022healthcare}. 

In general, taxonomies often represent hierarchical structures of concepts, while ontologies focus more on the structure and relationships of the network \cite{Marques2024}. As organizations seek to unify disparate classification systems, for instance, to integrate medical code consistency \cite{Tiina}, better synthesize e-commerce inventories \cite{Shi}, or consolidate educational frameworks or objectives \cite{Mohammad}, the need for accurate, efficient, and scalable content alignment and mapping is critical. The siloed structure of most taxonomies—a media company, for instance, might use different content categorization systems across its print, digital, and video platforms—impedes cross-functional insights and knowledge reuse across organizations. As organizations seek to leverage their collective knowledge, integrating these disparate taxonomies into an unified ontology has become both critical and challenging. 



Taxonomy alignment, the process of mapping between concept systems, traditionally requires time-consuming manual review by experts. Taxonomy alignment seeks to identify semantic correspondences (e.g., equivalences, subsumptions) between different ontologies or taxonomies \cite{Euzenat2013}. These semantic correspondences are critical for enabling interoperability between heterogeneous systems, facilitating data integration, and supporting knowledge-based applications. 

Although early alignment efforts have historically relied on manual curation efforts, often by domain experts, they have produced relatively accurate matches, though at small scales \cite{Noy2000}. However, not surprisingly, as ontologies increased in size, complexity, and domain coverage, manual approaches have become prohibitively more expensive and time-consuming \cite{Fahad}. 

Conventional machine learning approaches require extensive labeled datasets yet struggle with the nuanced contextual reasoning these alignment tasks demand. In contrast, large language model (LLM)-based approaches, such as ours, work with limited labeled data and captures complex semantic relationships. Additionally, LLMs generate detailed rationales, which accelerate expert validation.

In this paper, we present an automated framework for taxonomy alignment that leverages LLMs to map concepts across taxonomies in different domains and functions. Building on a foundation of expert calibration, systematic prompt refinement, and human validation, our approach balances automation with expert judgment to achieve high-quality ontology mappings at scale. We validate our framework in a domain-specific case study where the goal is to determine whether one concept is essential to the definition or realization of another. 

\section{Related Work}

\subsection{Historical Machine Learning (ML) Methods}
To address the need for cheaper and less time-consuming efforts, semi-automated systems were created, such as PROMPT \cite{Noy2000} and GLUE \cite{Doan2003}. 
PROMPT was designed to assist in ontology merging by suggesting potential mappings as well as conflicts; this allowed human users to confirm or revise the alignments, and while it successfully reduced manual overhead in certain contexts, it relied heavily on structural heuristics and domain-specific rules, which limited its generalizability across different ontological frameworks \cite{Noy2000}. GLUE advanced this by incorporating machine learning techniques to probabilistically match concepts based on a combination of syntactic, semantic, and instance-based similarity metrics \cite{Doan2003}; this technique brought the systems a step closer towards full automation. Ultimately, these early systems introduced rule-based or machine learning techniques to reduce expert labor, but both systems required extensive domain tuning and often struggled with heterogeneous structures and complex relationships.

Since the creation of semi-automated systems like PROMPT and GLUE, more recent research has focused on developing machine learning models that can generalize across domains and better capture semantic relationships. For instance, unsupervised and embedding-based approaches, like word2vec or GloVe, have been used to represent ontology terms in continuous vector spaces, enabling similarity computations based on distributional semantics \cite{Mikolov}\cite{Pennington}. A few years later, building upon the foundation set by unsupervised and embedding-based approaches, neural networks (GNNs) and knowledge graph embeddings were created to encode lexical and structural elements of ontologies for alignment tasks \cite{Ye}. Despite these advances, many models still require large labeled datasets for training and often lack interpretability—making it difficult for domain experts to understand or validate alignment decisions \cite{Li}. 

As the field reaches this inflection point between traditional methods and newer approaches, \cite{neuhaus2023} offers a critical perspective on how ontologies might evolve in the era of large language models, highlighting both the opportunities and challenges at this intersection. This transition has prompted growing interest in the use and application of Large Language Models (LLMs) within the context of ontologies and taxonomies.

\subsection{Emergence of LLMs for Semantic Tasks and Reasoning} \label{sec:llms-semantic}
Recent advances in LLMs have transformed the possibilities of semantic processing and automated reasoning, due to the models' capacities for deep natural language understanding, contextual reasoning, and flexible knowledge representation \cite{Brown2020}. These multifacted abilities makes LLMs well-suited for ontology-specific tasks such as entity mapping and axiom generation, where historical ML approaches have often fallen short \cite{babaei2023llms4ol, valcalvo2024ontogenix}. 

Early work by \cite{rajani2019} demonstrated that language models can generate explanations for their reasoning, establishing a foundation for interpretable AI. Building on this, researchers developed various prompting techniques to elicit reasoning capabilities from LLMs. Chain-of-thought prompting \cite{wei2022chain} and other reasoning frameworks \cite{creswell2022selection} showed promise in producing both structured reasoning and improved outputs, while \cite{Kojima2022} revealed that LLMs possess inherent zero-shot reasoning abilities that can be activated through carefully designed prompts. Beyond manual prompt engineering, automated optimization frameworks have emerged as systematic alternatives. AutoPrompt \cite{shin2020autoprompt} pioneered gradient-based search for optimal prompts, while later approaches expanded these capabilities: APE \cite{zhou2023large} uses LLMs themselves as prompt engineers; OPRO \cite{yang2024large} frames prompt optimization as a reinforcement learning problem; EvoPrompt \cite{guo2024connecting} applies evolutionary algorithms; and MIPRO \cite{OpsahlOng2024} employs Bayesian optimization for multi-stage prompt refinement. These approaches systematically generate and refine prompt instructions, often outperforming human-designed instructions for complex tasks. Despite these advances, LLM outputs may remain inconsistent if prompts are not meticulously designed \cite{raj2024improving}. Recent research emphasizes that scaling beyond typical few-shot settings substantially improves performance. \cite{wang2023context} demonstrate that many-shot prompting with tens or hundreds of examples can significantly boost in-context learning capabilities, especially for complex tasks requiring nuanced understanding of semantic relationships.

\subsection{LLMs versus Traditional ML in Classification}
The landscape of text classification reveals a spectrum of approaches with distinct trade-offs. Traditional machine learning methods such as Support Vector Machines (SVMs) and Naive Bayes classifiers represent the conventional end of this spectrum, relying on engineered features and performing well with sufficient annotated data \cite{sebastiani2001machine, wang2012baselines}. Moving toward greater complexity, fine-tuned encoder models like BERT occupy a middle ground, leveraging contextual embeddings to capture linguistic patterns beyond the reach of traditional methods \cite{qiu2024chatgpt}. 

At the far end stand large language models, which, despite their computational demands, demonstrate superior capacity for understanding nuanced, context-dependent relationships crucial for tasks like ontology alignment \cite{Chako}. Recent research reinforces this advantage, showing that strategically prompted LLMs—especially those utilizing few- or many-shot demonstrations—consistently outperform both traditional algorithms and smaller neural models on complex classification challenges \cite{wang2023context, suzgun2022challenging}. This progression illustrates not merely differences in technical approach but a fundamental evolution in how machines process and interpret textual information.

\subsection{Related Work Summary and Study Significance}
Ontologies and taxonomies are vital tools for organizing knowledge across domains such as healthcare, education, and commerce \cite{Osman2022}. As these systems scale in complexity, manual alignment methods have become too costly and inconsistent \cite{Noy2000, Fahad}.

Early semi-automated tools like PROMPT and GLUE reduced expert labor through rule-based and ML techniques but required extensive tuning and struggled with diverse ontologies \cite{Noy2000, Doan2003}. Later models using embeddings and neural networks (e.g., word2vec, GloVe, GNNs) improved semantic understanding but often lacked transparency and required large labeled datasets \cite{Mikolov, Pennington, Ye, Li}. However, LLMs offer a more flexible, data-efficient alternative, showing strong performance on ontology-related tasks with minimal supervision \cite{babaei2023llms4ol, valcalvo2024ontogenix}. Prompt engineering and optimization techniques—such as few-shot learning, many-shot prompting, and frameworks like MIPRO, OPRO, and APE—further enhance accuracy and adaptability without extensive retraining \cite{Brown2020, wei2022chain, wang2023context, OpsahlOng2024, zhou2023large, yang2024large}.

Building on these research insights, our approach integrates domain calibration with an LLM-based alignment pipeline, striking a balance between automation and expert validation. Our framework integrates systematic prompt refinement with minimal example-based guidance, enabling the model to produce both classification outcomes and rationales without requiring extensive labeled datasets. 

We employ both manual optimization techniques and automated approaches like MIPRO to maximize performance while maintaining transparency in the reasoning process. Overall, our work addresses scalability and complexity challenges by leveraging LLMs to generate high-quality alignments while retaining expert oversight for domain-specific edge cases.

\section{Methodology}

\subsection{Data Annotations}

\subsubsection{Initial Annotation Phase}
We began with our initial annotation phase where a set of 973 concept pairs was independently labeled by four annotators. Each instance is represented as a pair of textual descriptions: $(\text{Concept A},\ \text{Concept B})$. The annotation task assessed whether Concept A is essential for the realization of Concept B. The output was a binary label: \textit{Required} (indicating Concept A is essential for Concept B's completion) or \textit{Not Required} (indicating Concept A may be valuable but is not essential for Concept B's completion). Early observations revealed notable disagreements among annotators, with only 22\% of concept pairs achieving unanimous agreement.

\subsubsection{Calibration Sessions} \label{para:calibration}
To address low agreement among annotators, we conducted calibration sessions for the 759 non-unanimous pairs (78\% of total). Initially, the annotators used a significance-based Likert scale with categories (\textit{Critical}, \textit{Significant}, \textit{Beneficial}, \textit{Marginal}, and \textit{Irrelevant}) to rate each instance. However, the subjective interpretation of these terms led to varied assessments. During calibration discussions, we introduced an alternative frequency-based scale with categories (\textit{Always}, \textit{Usually}, \textit{Often}, \textit{Sometimes}, and \textit{Not Necessary}). Under this scheme, only instances rated as \textit{Always} were mapped to the label \textit{Required}, while all other ratings were considered \textit{Not Required}. 

For example, "verbal communication" was determined to be \textit{Always Necessary} (Required) for "mentoring" since it inherently involves spoken interaction, while "written communication" was classified as \textit{Sometimes Necessary} (Not Required) as mentoring activities rarely depend solely on written exchanges.

\subsubsection{Annotation Results} \label{para:annotation}
The calibration process established a ground truth dataset, resulting in 314 linkage rationales. The final distribution showed 34\% of concepts marked as \textit{Required} and 66\% as \textit{Not Required}. Comparing initial independent annotations against the calibrated ground truth across the full dataset of 973 samples yielded metrics of 0.69 for precision, recall, and accuracy, with a 0.68 F1-score (Table \ref{tab:annotations}), highlighting both the challenges in the annotation process and the importance of calibration. Unlike later model-based experiments that use data partitioning, these metrics represent the initial human benchmark on the entire collection. On average, reaching a consensus for each instance took approximately 4 minutes, with the overall calibration process spanning about 50 hours for the entire 973 samples.

\begin{table}[h]
\centering
\begin{tabular}{cccc}
\hline
Precision & Recall & F1 Score & Accuracy \\
\hline
0.690 & 0.693 & 0.682 & 0.691 \\
\hline
\end{tabular}
\caption{Initial vs Calibrated Human Annotations (Full Dataset of 973 Samples)}
\label{tab:annotations}
\end{table}

\subsection{Prompt Optimization}

\subsubsection{Manual Instruction Optimization (Zero-shot Prompting)}
With a calibrated ground truth dataset established, we focused on developing an LLM-based approach to scale this classification task. Building on the discovery that LLMs exhibit strong zero-shot reasoning abilities \cite{Kojima2022}, our initial experiments focused on designing effective zero-shot instructions to guide model reasoning without exemplars. Transitioning from human annotation to an LLM-based system required thoughtful prompt engineering to guide the model toward decision patterns consistent with our calibration sessions. Our prompt optimization strategy evolved through several stages, beginning with basic zero-shot instructions and progressively incorporating insights from the calibration process.

We first manually refined the prompt by integrating detailed guidelines and reasoning processes derived from our calibration sessions. Specifically, we elaborated on the criteria for assessing concept necessity, provided illustrative examples to clarify ambiguous cases, and incorporated the documented rationales from annotators. These enhancements enriched the prompt with explicit context and decision-making cues, thereby improving the LLM's understanding and performance. (Detailed prompt formulations are available in Appendix~\ref{app:human_llm_instructions}.)

\subsubsection{Automated Instruction Optimization (Zero-shot Prompting)}
To build upon our manually refined prompt, we leveraged automated prompt optimization techniques. Drawing from the frameworks discussed in Section~\ref{sec:llms-semantic}, we employed the Multiprompt Instruction Proposal Optimizer (MIPRO) \cite{OpsahlOng2024}, for its effectiveness with structured reasoning tasks and ability to generate contextually relevant instructions.

MIPRO systematically enhances prompts in two phases: proposal generation and credit assignment. In the proposal generation phase, it creates diverse candidate prompts by bootstrapping few-shot examples from the training data. It leverages multiple contextual sources—such as training dataset summaries, prompt summaries, previously bootstrapped examples, and randomly sampled generation tips—to produce new candidate instructions that capture the nuanced requirements of our task.

In the credit assignment phase, MIPRO applies Bayesian optimization to evaluate these candidate prompts. A Bayesian model estimates each prompt's performance using evaluation metrics (e.g., accuracy or exact match) on validation mini-batches. The optimizer then iteratively refines and selects the most effective combinations of candidate instructions and demonstrations in the form of example inputs and outputs, periodically validating on the full set.

We applied MIPRO to a dataset of 963 annotated samples, partitioned equally for training (instruction generation), development (credit assignment with exact match evaluation), and testing. Using zero-shot demonstrations with a mini-batch size of 25 and full validations every 10 mini-batch trials, this process generated optimized prompts.

\subsubsection{Human-Generated Rationales (Few-shot Prompting)}
To further enhance the LLM's performance, we incorporated human-generated rationales into the few-shot prompts. During the calibration sessions described in Section~\ref{para:calibration}, annotators resolved disagreements and documented 314 rationales that captured the reasoning behind each labeling decision (see Appendix~\ref{app:human_llm_rationales} for an example of human-generated rationales). These rationales were integrated into the prompts alongside the optimized instructions from the previous section. We conducted experiments using both 3-shot and 10-shot demonstration settings, with each demonstration including a pair of textual descriptions, the calibrated binary classification, and the associated rationale.

\subsubsection{LLM-Generated Rationales (Few-shot Prompting)}
Building on our experiments with human-generated rationales, we explored the use of LLM-generated rationales as an alternative. Recent research indicates that model-generated chain-of-thought rationales can surpass human-written explanations \cite{wang2023selfconsistency, madaan2022language}. Our goal was to optimize these LLM-generated rationales for concept-linkage tasks and compare their effectiveness with human-generated ones. Using the same dataset split as in the previous section, we leveraged MIPRO to generate and refine candidate rationales. Specifically, MIPRO produced six candidate rationales per demonstration, and validation testing was used to select the most effective versions. We evaluated this approach using both 3-shot and 10-shot demonstration settings. (See Appendix~\ref{app:human_llm_rationales} for examples of LLM-generated rationales.)

\subsubsection{Many-Shot Demonstrations (Many-shot Prompting)}
While our few-shot demonstrations provided insights, recent work by \cite{wang2023context} suggests that scaling the number of demonstrations can lead to performance improvements through in-context learning (ICL) \cite{Brown2020}. Motivated by these findings, we extended our few-shot approach to the many-shot regime, hypothesizing that an increased number of examples would enhance LLM performance by exposing the model to diverse reasoning patterns and enabling pattern recognition across demonstrations. To this end, we created a demonstration pool by generating LLM rationales for 642 concept pairs from our training and development datasets. Each entry in the pool comprised three components: the pair of textual descriptions, the LLM-generated rationale, and the ground-truth binary classification.

For evaluation, we constructed prompts by randomly selecting between 50 and 300 demonstrations from this pool to assess how increasing the context window affects the model’s ability to generate accurate rationales and classifications for new, unseen test samples.

\section{Experiments}
\subsection{Experimental Setup}
Our experiments were conducted on a calibrated dataset of 973 annotated concept pairs, where each instance consists of two textual descriptions of concepts, along with a binary label (\textit{Required}, \textit{Not Required}) indicating whether one concept is essential to the definition or realization of the other. The dataset was partitioned into training, development, and test sets as described previously.

We evaluated our framework using several LLMs, including Anthropic's Claude 3.7 Sonnet v1, Claude 3.5 Sonnet v2, and Claude 3 Haiku v1. In our assessment of Claude 3.7 Sonnet, we examined both standard inference mode (without thinking) and enhanced thinking mode (with 10,000 reasoning tokens). Reasoning token budget optimization is explored in Section \ref{sec:adaptive_token_budgeting} as future work. Additionally, we conducted prompt optimization experiments across zero-shot, few-shot, and many-shot demonstration settings to determine optimal prompting strategies.

Performance was evaluated by comparing the ground truth and generated labels (\textit{Required}, \textit{Not Required}) using precision, recall, and accuracy, with emphasis on the F1-score as our primary evaluation criterion. Human annotation performance established a benchmark F1-score of 0.68 for comparison. All experiments were executed in an offline environment where inference latency was not considered a critical constraint for evaluation purposes.

\subsection{Results and Analysis}

\subsubsection{Prompt Optimization}
\begin{figure*}[h]
    \centering
    \includegraphics[width=\textwidth]{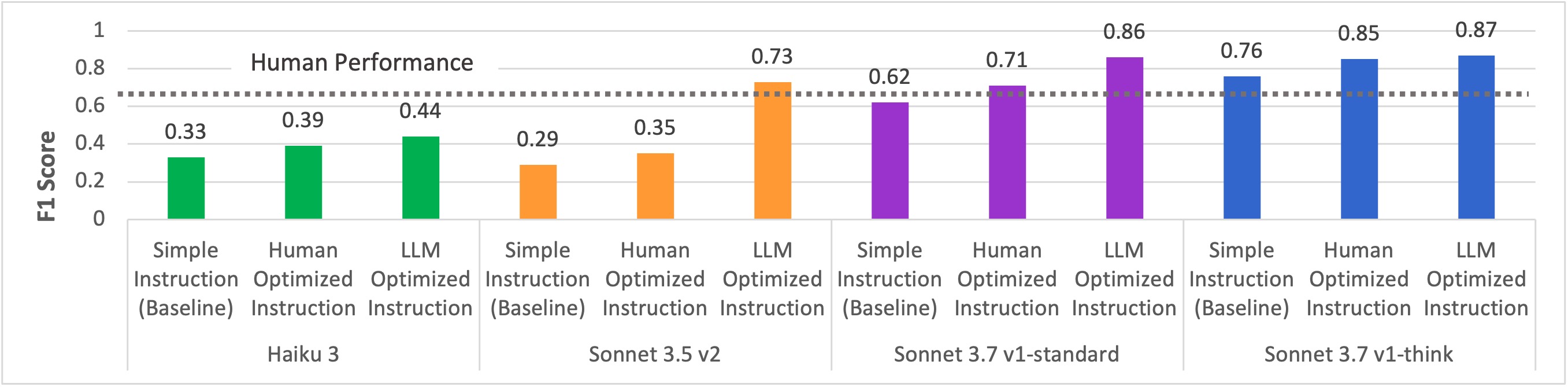}
    \caption{Performance of Human vs LLM Optimized Instructions (Zero-shot)}
    \label{fig:human_vs_llm_zeroshot}
\end{figure*}

\paragraph{Manual Instruction Optimization (Zero-shot Prompting)} 
Using a baseline one-sentence instruction, models showed varying F1 scores (Haiku3: 0.33, Sonnet3.5: 0.29, Sonnet3.7-standard: 0.62, Sonnet3.7-think: 0.76). By enhancing the prompt with guidelines and reasoning processes from our calibration sessions, we observed performance improvements across models (Haiku: 0.39, Sonnet3.5: 0.35, Sonnet3.7-standard: 0.71, Sonnet3.7-think: 0.85). As Figure~\ref{fig:human_vs_llm_zeroshot} illustrates, the benefits of prompt optimization scale with model capacity, with larger models showing greater receptiveness to enhanced instructions. Notably, both Sonnet3.7-standard and Sonnet3.7-think exceeded human performance (Human: 0.68). See Appendices \ref{app:human_llm_instructions} and \ref{app:instruction_performance} for prompts and full results.

\paragraph{Automated Instruction Optimization via MIPRO (Zero-shot Prompting)}
We applied the MIPRO framework to transform a single-sentence instruction into a structured prompt of approximately 40 sentences (see Appendix~\ref{app:human_llm_instructions} for an example prompt). The enhanced prompt integrated explicit task guidelines, reasoning steps, and edge case handling. This optimization improved performance in F1 across all models (Haiku: 0.39 to 0.44, Sonnet3.5: 0.35 to 0.73, Sonnet3.7-standard: 0.71 to 0.86, and Sonnet 3.7-think: 0.85 to 0.87). As shown in Figure~\ref{fig:human_vs_llm_zeroshot}, Sonnet 3.5, and Sonnet 3.7-standard exceeded human benchmark (Human: 0.68) after optimization. 

\begin{figure}[h]
    \centering
    \includegraphics[width=\columnwidth]{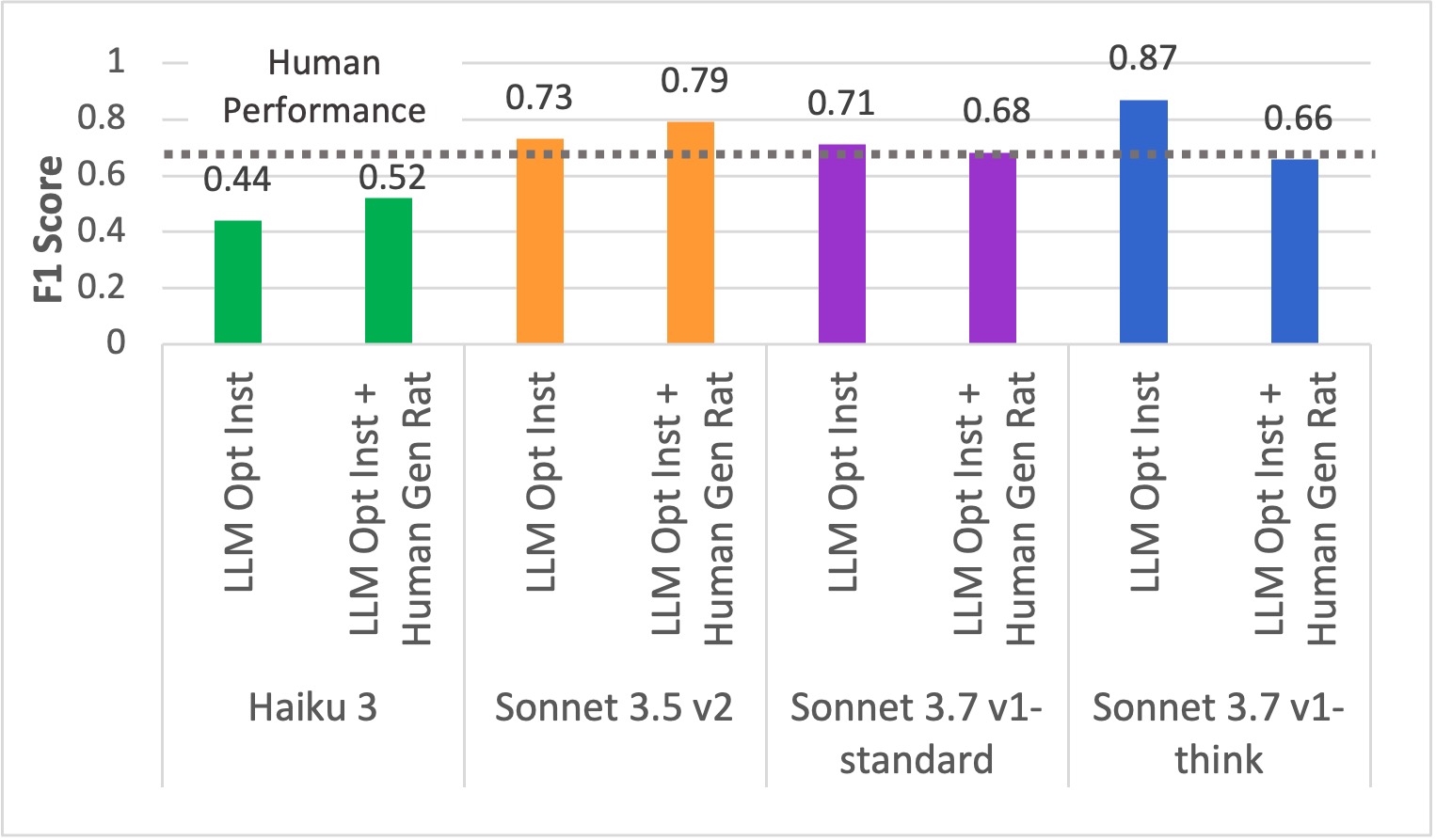}
    \caption{Performance comparison between LLM-optimized instruction alone versus LLM-optimized instruction supplemented with human-generated rationales (10-shot prompting). Note: LLM Opt Inst = LLM-optimized Instruction, Human Gen Rat = Human-generated Rationales.}
    \label{fig:llm_instruction_plus_human_rationales}
\end{figure}

\paragraph{Human-Generated Rationales (Few-shot Prompting)}
Following instruction optimization, we investigated the efficacy of augmenting the instructions with human-generated rationales. While MIPRO provided methodological guidance, human rationales captured expert reasoning patterns documented during calibration sessions for each pair (see Section \ref{para:annotation} for calibration details and Appendix \ref{app:human_llm_rationales} for an example of human-generated rationales). We examined the combination of LLM-optimized instructions with human-generated rationales. As illustrated in Figure~\ref{fig:llm_instruction_plus_human_rationales}, incorporating 10 human-generated rationales as demonstrations yielded performance improvements for Haiku 3 and Sonnet 3.5, but not for Sonnet 3.7. Specifically, Haiku 3's F1 score improved from 0.44 (LLM-optimized instruction only) to 0.52 (LLM-optimized instruction with human-generated rationales), while Sonnet 3.5 improved from 0.73 to 0.79. 

\begin{figure*}[h]
\centering
\includegraphics[width=\textwidth]{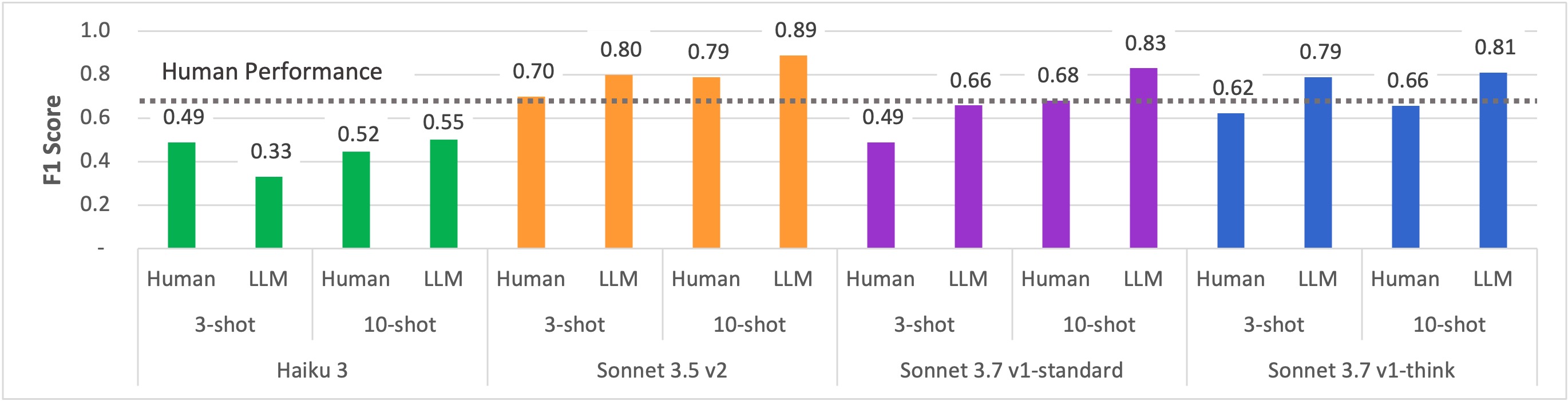}
\caption{Performance comparison between human-generated and LLM-generated rationales across different models (few-shot prompting)}
\label{fig:human_vs_llm_fewshot}
\end{figure*}

These empirical results suggest a complementary relationship between optimized instructions and exemplar reasoning, where their combination produces performance gains exceeding those observed from either approach in isolation.

\paragraph{LLM-Generated Rationales (Few-shot Prompting)}
Building on experiments with human-generated rationales, we investigated whether LLM-generated rationales could serve as an alternative. As Figure~\ref{fig:human_vs_llm_fewshot} demonstrates, these rationales outperformed human-generated ones across most configurations. For most models in both 3-shot and 10-shot settings, LLM-generated rationales yielded superior results compared to human-generated rationales, with the exception being the 3-shot setting with Haiku 3. The largest improvements occurred with Sonnet 3.7, where both standard and think modes exhibited an F1 score increase of 0.17 when using LLM-generated rationales.

\begin{figure}[h]
    \centering
    \includegraphics[width=\columnwidth]{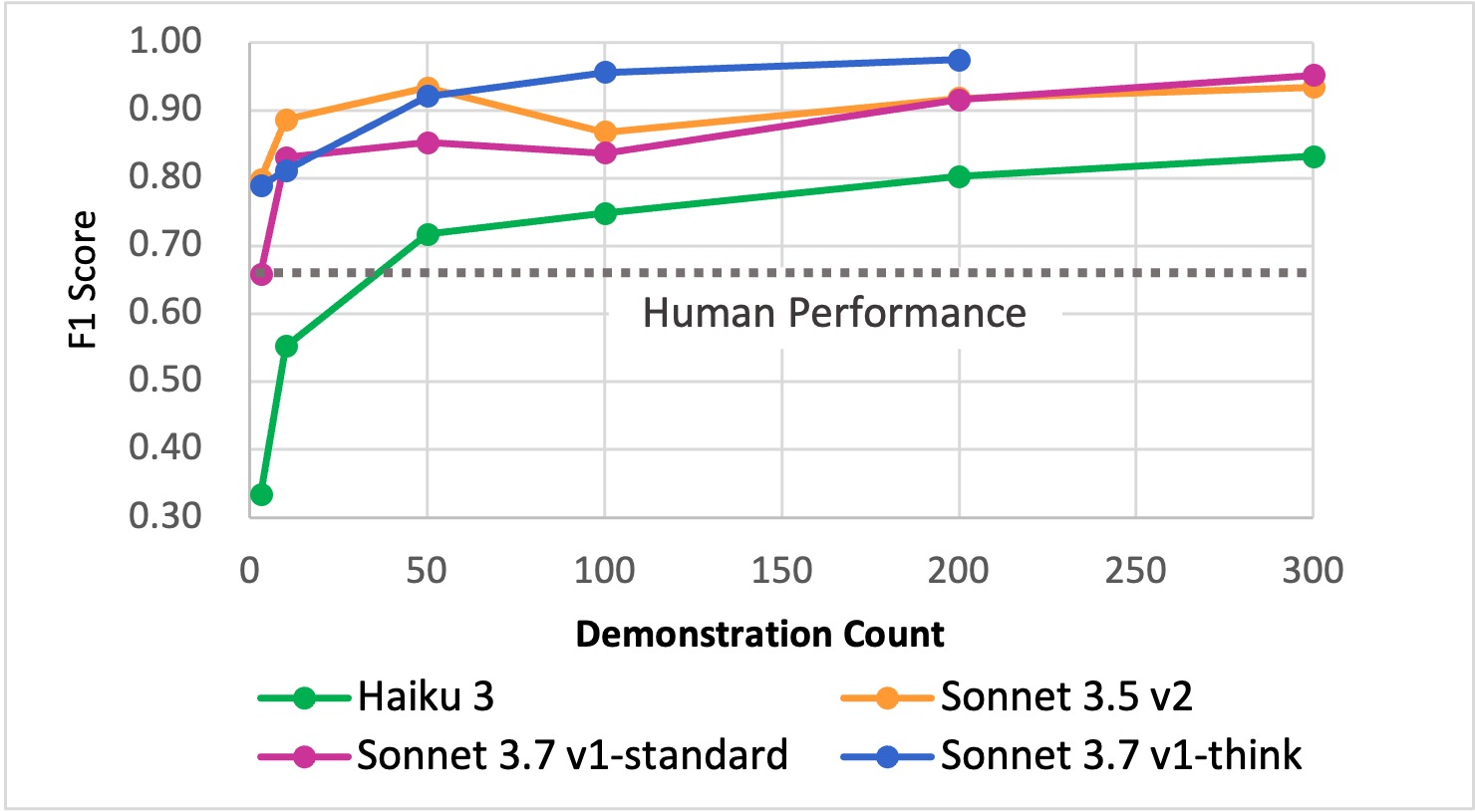}
    \caption{Performance of Many-Shot Demonstrations}
    \label{fig:performance_manyshot_demo}
\end{figure}

\paragraph{Many-Shot Demonstrations}
Scaling the number of demonstrations up to 300 further boosted model performance across models. Figure~\ref{fig:performance_manyshot_demo} demonstrates that at 50 demonstrations, all models, including Haiku 3 (the smallest model tested), exceeded the human benchmark. 

Performance improved as demonstrations increased from 3 to 300, with model-specific patterns. Haiku 3's F1 scores monotonically increased from 0.33 to 0.83, showing the gains throughout the scaling range. The Sonnet family demonstrated higher performance but with diminishing returns beyond 50-100 examples. Sonnet 3.7-standard reached 0.95 with 300 demonstrations, while Sonnet 3.7-think achieved 0.97 at 200 demonstrations. With 10,000 tokens already devoted to reasoning processes, the think variant reached its context limit at 200 demonstrations, preventing tests at 300 demonstrations—a constraint that motivates our exploration of adaptive budgeting in Section~\ref{sec:adaptive_token_budgeting}.

Results indicate that scaling benefits vary with model capacity. Smaller models continue to benefit from additional examples beyond where larger models plateau, suggesting n-shot prompting approaches should be tailored to specific model architectures. See Appendix \ref{app:performance_rationales} for full results.

\section{Discussion}
\subsection{Human vs. Automated Instruction Optimization}
Our experiment comparing instruction optimization strategies revealed differences between manual and automated approaches. Through manual optimization, we expanded a one-line prompt to approximately 10 lines by incorporating guidelines and reasoning steps documented during calibration sessions, which yielded performance improvements across all models. Haiku3 and Sonnet 3.5 showed modest F1 improvements (from 0.33 to 0.39 and 0.29 to 0.35 respectively), while Sonnet 3.7 variants demonstrated more substantial gains (standard mode: from 0.62 to 0.71, think mode: from 0.76 to 0.85). These results confirm that extended instructions enhance model performance, though the improvement varies according to model capabilities.

In contrast, the MIPRO automated optimization technique transformed the original one-line prompt into a detailed set of instructions spanning approximately 40 lines. This approach systematically captured reasoning patterns and addressed edge cases that might not have been identified through manual refinement alone. With automated optimization, Sonnet 3.5 demonstrated the largest F1 improvement of 0.44, while Sonnet 3.7-standard improved by 0.24 and Sonnet 3.7-think by 0.02, highlighting varying responsiveness to instruction optimization across model architectures. Notably, after automated optimization, all Sonnet family models exceeded the human benchmark of 0.68. Results indicate that enhanced instruction detail contributes to performance gains, with effects contingent on model architecture and baseline capabilities. See Appendix \ref{app:human_llm_instructions} for examples of manual and automated optimized prompts.

\subsection{Human-Generated vs. LLM-Generated Rationales}
Incorporating rationales into optimized instructions consistently enhanced model performance across experiments. While human-generated rationales showed variable impact—Haiku 3 gained 18\% in F1 score but Sonnet 3.7 v1-think declined by 24\% in the 10-shot setting (Figure~\ref{fig:llm_instruction_plus_human_rationales}). Whereas LLM-generated rationales delivered superior and more reliable improvements (Figure~\ref{fig:human_vs_llm_fewshot}). The performance advantage was substantial: Sonnet 3.7 v1-standard achieved a 34\% improvement with automated optimization compared to only 15\% with human-optimized prompts (see Appendix~\ref{app:instruction_performance} for the full results).

The success of LLM-generated rationales stems from their systematic framework combining component-level assessment, necessity evaluation, and counterfactual reasoning—a structure particularly effective for resolving edge cases and ambiguous relationships. These results challenge conventional wisdom about human expertise as the upper bound for instructional quality, demonstrating that algorithmic reasoning patterns outperform human explanations in guiding complex classification decisions.

\subsection{Zero-shot, Few-shot, vs. Many-shot Demonstrations}
The progression from zero-shot to many-shot demonstration settings reveals a clear performance pattern. Zero-shot prompting with baseline instructions established a performance baseline, while few-shot demonstrations incorporating either human or LLM rationales yielded measurable improvements (Figure~\ref{fig:performance_manyshot_demo}). The most substantial gains emerged in many-shot settings, where scaling to 300 examples enabled all tested models to surpass the human F1-score benchmark of 0.68. Notably, Sonnet 3.7 v1-think reached a score of 0.97, approaching near-perfect classification accuracy.

Exposure to diverse reasoning patterns through additional examples enhanced models' ability to resolve ambiguities and generalize to unseen cases. As the number of demonstrations increased, the performance gap between models narrowed, indicating that lower-capacity models derive proportionally greater benefits from additional examples. While these results establish a new ceiling for concept linkage classification, they present a practical trade-off: many-shot approaches deliver superior performance but require longer prompts, increasing latency and inference costs. These findings suggest that n-shot learning strategies should be calibrated to balance model capabilities against computational constraints in deployment scenarios.

\subsection{Human-in-the-Loop Validation and Operationalization Strategy}
Our LLM approach significantly outperformed the human benchmark (F1 score of 0.97 compared to 0.68). This performance gap raises the question: When LLM predictions diverge from post-calibrated human ground truth annotations, which source is more reliable? We hypothesized that the LLM might be identifying oversights in the post-calibrated human annotations rather than making errors. This section evaluates this hypothesis through systematic analysis of discrepancies between LLM and human annotations. We also consider the implications for prioritizing expert review in real-world scenarios, where full manual verification becomes infeasible as taxonomies scale in size.

\begin{figure}[h]
    \centering
    \includegraphics[width=\columnwidth]{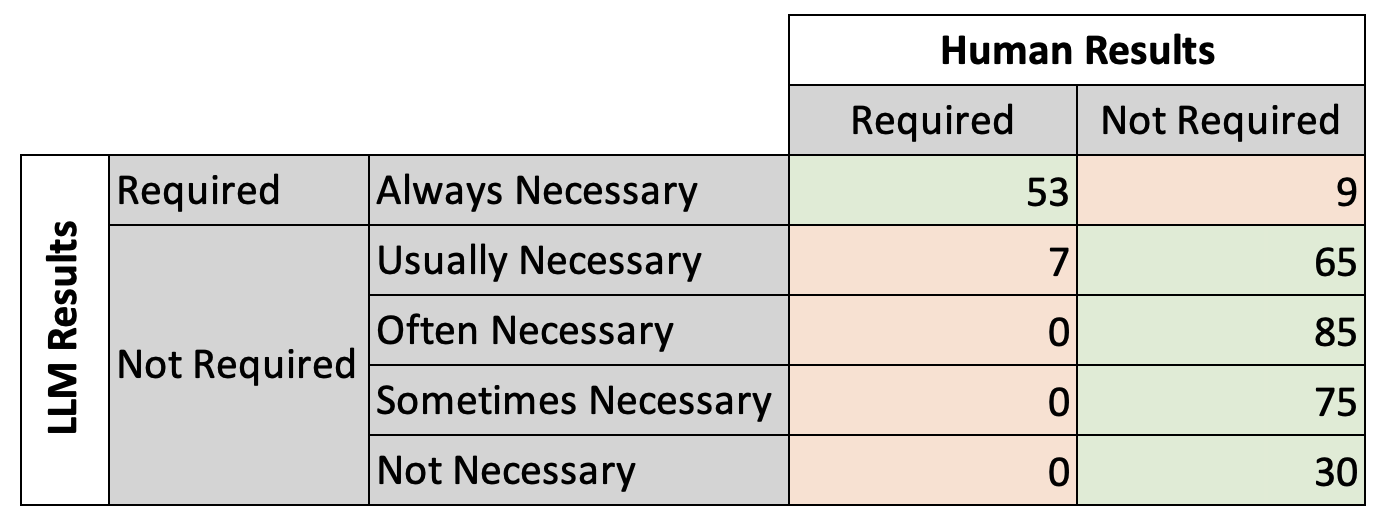}
    \caption{Confusion Matrix: LLM vs Human Results}
    \label{tab:confusion-matrix}
\end{figure}

To investigate these discrepancies systematically, we analyzed instances where the LLM and post-calibrated ground truth disagreed. Table \ref{tab:confusion-matrix} presents the distribution of these disagreements. We identified 9 cases where the LLM classified concepts as \textit{Required} ("Always Necessary") while human annotations marked them \textit{Not Required}, and 7 cases where the LLM classified concepts as \textit{Not Required} ("Usually Necessary") while human annotations marked them \textit{Required}.

Human annotators from the original calibration sessions reviewed these 16 cases alongside the LLM's detailed reasoning. This review process revealed that in all examined cases, the LLM classifications were superior—all 9 cases were confirmed to be false negatives and all 7 cases were false positives in the human annotations. These findings confirmed our hypothesis that the LLM had identified legitimate oversights in the human annotation process.

The inclusion of LLM-generated rationales improved review efficiency. Whereas the initial calibration sessions described in Section \ref{para:calibration} required four-minute discussions to reach consensus, annotators reached immediate agreement after reading the LLM rationales, reducing the review time to about one minute per instance—primarily spent on reading.

These results have implications for ontology construction and maintenance. First, they demonstrate that even carefully calibrated human annotations can contain errors that LLMs with appropriate prompt engineering can detect. Second, they highlight the value of an iterative human-in-the-loop process where LLM outputs refine both model performance and training data quality.

For organizations maintaining ontologies at scale, our findings suggest an optimal operational strategy: combine automated LLM processing with targeted expert oversight. High-confidence LLM predictions can be accepted automatically, while ambiguous cases are flagged for expert review. This approach creates an efficient feedback loop that enhances both accuracy and scalability while deploying human expertise where it adds the most value.

\subsection{Practical Application}

Organizations develop taxonomies across functions that often evolve into silos impeding organizational effectiveness. In workforce management, these disconnected knowledge structures contribute to skills gaps and talent utilization challenges—with the World Economic Forum estimating over 100 million people across 18 economies are underutilizing their existing skills \cite{di2023future}, thus hindering innovation and industry transformation.
 
To address this challenge, we developed a unified ontology by applying our LLM-based framework with custom SKOS extensions \cite{miles2009skos} that implement specialized mapping properties (\textit{myskos:isRequiredFor} and \textit{myskos:isNotRequiredFor}) to define prerequisite relationships between skills. By linking previously disparate taxonomies, we've created an integrated skills library that enables organizations to effectively upskill and reskill their workforce. This comprehensive solution powers talent management products while providing a knowledge base for LLM models that support learning initiatives, performance management, and talent assessment across organizations.

\section{Future Work}
\subsection{Dynamic Example Selection}
Although many-shot demonstrations significantly improved performance, they introduced computational overhead by requiring models to process hundreds of examples per inference. Future work should explore dynamic example selection strategies that identify relevant demonstrations based on semantic similarity to the query. This approach may reduce the number of examples needed from hundreds to dozens while maintaining performance levels. Such optimization could decrease token consumption, latency, and inference costs, making LLM deployment more practical in resource-constrained environments. 
Moreover, the use of LLMs as judges could be explored to evaluate and arbitrate annotation disagreements, a method investigated by \cite{zheng2023judging, bedemariam2025potential}, as a potential scalable alternative to traditional human-centric evaluation. This approach may offer efficiency gains, though careful evaluation would be needed to ensure it can capture the subtle, context-specific nuances often found in qualitative data \cite{bhaduri2024reconciling}.

\subsection{Adaptive Reasoning Token Budgeting}
\label{sec:adaptive_token_budgeting}
While our current framework utilizes fixed reasoning tokens of 10,000, recent research suggests that dynamically adjusting token budgets based on task complexity can enhance efficiency and performance. Han and Johnson (2024) introduced TALE, a token-budget-aware LLM reasoning framework that estimates optimal token allocations for varying problem types, reducing token consumption while maintaining accuracy \cite{han2024token}. Similarly, Lee and Chen (2025) investigated the trade-off between reasoning length and accuracy, demonstrating that each task exhibits a 'token complexity'—the minimum number of tokens necessary for successful problem resolution \cite{lee2025token}. Incorporating adaptive token budgeting into our framework could optimize resource utilization and improve scalability, particularly for large-scale ontology alignment challenges.

\section{Conclusion}
We proposed a novel framework leveraging large language models to automate ontology alignment tasks, significantly outperforming traditional manual approaches. By combining expert calibration, systematic prompt optimization, and human-in-the-loop validation, our method achieved classification performance well beyond human benchmarks while reducing the time and effort required for concept mapping. Experimental results demonstrated that automated prompt engineering techniques, LLM-generated rationales, and many-shot demonstrations all contributed to substantial performance improvements across model architectures.

Our findings reveal that LLM-generated rationales consistently outperform human-authored explanations. While human experts achieved only 22\% unanimous agreement and an F1-score of 0.68, our LLM approach reached an impressive 0.97 F1-score. This performance advantage was substantial: Sonnet 3.7 v1-standard achieved a 34\% improvement with automated optimization compared to only 15\% with human-optimized prompts. The success of LLM-generated rationales stems from their systematic framework combining component-level assessment, necessity evaluation, and counterfactual reasoning.

The human-in-the-loop validation process further demonstrated the model's effectiveness. When reviewing disagreements between LLM and human annotations, experts confirmed that human annotators had overlooked linkages in all examined cases (9 false negatives and 7 false positives), highlighting the value of an iterative refinement process. These findings show that mapping relationships between taxonomies connects isolated systems, unifying fragmented information. The proposed framework strikes an effective balance between automation and expert oversight—allowing high-confidence cases to be processed automatically while flagging ambiguous instances for human review. This approach creates a scalable, consistent system that transforms taxonomy alignment from a labor-intensive process into an automated workflow with better quality, enabling organizations to build practical applications atop unified knowledge structures.

\section*{Acknowledgments}
\begin{small}
We extend our sincere gratitude to Jason VanDuine, Henry Zhu, and Hideo Kobayashi for their insightful feedback and support that strengthened this work.
\end{small}


\clearpage
\onecolumn
\appendices

\FloatBarrier 
\section{Human and LLM Optimized Instructions}
\label{app:human_llm_instructions}
A comparison of three instruction approaches shows progression from a simple baseline instruction to a human-optimized version with structured evaluation steps, culminating in an LLM-optimized instruction that provides detailed guidance on competency analysis, reasoning steps, necessity ratings, and final determination criteria.

\begin{center}
\captionof{table}{Comparison of Instruction Types}
\label{tab:instruction_comparison}
\begin{tabular}{|p{0.2\textwidth}|p{0.35\textwidth}|p{0.45\textwidth}|}
\hline
\textbf{Simple Instruction} & \textbf{Human Optimized Instruction} & \textbf{LLM Optimized Instruction} \\
\hline
\begin{minipage}[t]{0.2\textwidth}
Identify if a competency is `Required` or `Not Required` to complete the responsibility. Output just `Required` or `Not Required`
\end{minipage} & 
\begin{minipage}[t]{0.35\textwidth}
As an ontological domain expert specializing in conceptual modeling, your role is to determine whether a specific concept is essential within a particular domain context.\\

1. Analyze the Relationship: Evaluate whether Concept A is essential to the definition or realization of Concept B. Provide your reasoning in the rationale section, ensuring your assessment strictly aligns with the conceptual definitions provided.\\

2. Rate Necessity: Using the rationale, rate the competency's necessity for performing the responsibility according to the Likert scale below:\\
-~Always Necessary: Essential and required in all circumstances.\\
-~Usually Necessary: Very important, with rare exceptions.\\
-~Often Necessary: Frequently helpful but not critical.\\
-~Sometimes Necessary: Occasionally useful.\\
-~Not Necessary: Never required.\\

3. Determine Requirement: Based on your rationale and Likert scale analysis, identify whether the competency is Required or Not Required for fulfilling the responsibility. Output only `Required` or `Not Required`.
\end{minipage} & 
\begin{minipage}[t]{0.45\textwidth}
You are an experienced ontological domain expert specializing in conceptual relationship analysis and knowledge mapping. Your task is to evaluate whether specific concepts are essential to other concepts by following this systematic approach:\\

1. First, carefully analyze:\\
-~Concept A and its definition\\
-~Concept B and its definition\\
-~How they fundamentally relate to each other\\

2. Provide detailed step-by-step reasoning that:\\
-~Breaks down the key components of Concept B\\
-~Examines how Concept A relates to these components\\
-~Considers if Concept B can be defined or realized without Concept A\\
-~Evaluates the impact of removing Concept A from the relationship\\

3. Rate the essentiality of Concept A to Concept B using this Likert scale:\\
-~Always Necessary: Essential and required in all circumstances\\
-~Usually Necessary: Very important with rare exceptions\\
-~Often Necessary: Frequently helpful but not critical\\
-~Sometimes Necessary: Occasionally useful\\
-~Not Necessary: Never required\\

4. Make a final determination:\\
-~Required: Concept B CANNOT be defined or realized without Concept A under ANY circumstances\\
-~Not Required: Concept A is helpful but not absolutely essential to Concept B\\

Important: Focus on whether Concept B can exist or be meaningfully defined at all without Concept A, not just whether it would be fully developed. A concept should only be marked as "Required" if its absence makes the other concept impossible to define or realize, not just less complete.\\

Provide your analysis, Likert rating, and final determination with clear supporting rationale.
\end{minipage} \\
\hline
\end{tabular}
\end{center}

\clearpage
\section{Human and LLM Generated Rationales}
\label{app:human_llm_rationales}
This table presents a comparison between two types of rationales—Human-authored and LLM-generated—that evaluate whether 'Verbal Communication' is a required competency for effectively mentoring another team member.

\begin{table}[h]
\centering
\caption{Comparison of Human and LLM Rationales}
\begin{tabular}{|p{0.4\textwidth}|p{0.6\textwidth}|}
\hline
\textbf{Human Rationale} & \textbf{LLM Rationale} \\
\hline
Mentoring another team member requires articulating expectations, sharing knowledge, and providing constructive feedback through effective verbal communication. & 
\begin{minipage}[t]{0.6\textwidth}
1. Let me analyze this case step by step:\\
- Break down the key components of mentorship:\\
- Knowledge transfer\\
- Skill development\\
- Performance feedback\\
- Career guidance\\
- Relationship building\\

2. Analyze the verbal communication requirements:\\
- Explaining complex concepts clearly\\
- Articulating expectations and goals\\
- Delivering constructive feedback\\
- Asking effective questions\\
- Active listening and responding\\

3. Evaluate the connection:\\
- Cannot transfer knowledge without clear verbal explanations\\
- Feedback delivery requires nuanced verbal communication\\
- Building mentor-mentee trust depends on verbal rapport\\
- Guidance becomes ineffective without articulate expression\\
- Misunderstandings multiply without verbal clarity\\

4. Consider if mentorship could succeed without verbal communication:\\
- Written communication alone lacks immediate clarification\\
- Non-verbal cues insufficient for complex knowledge transfer\\
- Time-sensitive feedback requires verbal exchange\\
- Relationship development hindered by lack of verbal interaction\\
- Impact assessment:\\

5. Failure in verbal communication would result in:\\
- Incomplete knowledge transfer\\
- Misinterpreted feedback\\
- Diminished mentee engagement\\
- Ineffective skill development\\
- Damaged mentor-mentee relationship\\

Answer: Required
\end{minipage} \\
\hline
\end{tabular}
\label{tab:rationale_comparison}
\end{table}

\clearpage
\FloatBarrier 
\section{Performance Comparison of Different Instructions}
\label{app:instruction_performance}
We evaluated various instruction strategies across multiple Anthropic models—Claude 3.7 Sonnet v1, Claude 3.5 Sonnet v2, and Claude 3 Haiku v1—comparing their performance using standard metrics: Precision, Recall, F1-score, and Accuracy. For Claude 3.7 Sonnet specifically, our analysis included both standard inference operation and enhanced thinking mode with 10,000 reasoning tokens. All metrics reported are weighted averages across classes, where weights are proportional to class frequencies in the dataset.

\begin{center}
\captionof{table}{Performance Comparison of Different Instructions (Weighted Average Metrics: Prec = Precision, Rec = Recall, Acc = Accuracy)}
\label{tab:instruction_performance_comparison}
\setlength{\tabcolsep}{3.5pt} 
\small 
\begin{tabular}{l*{16}{S[table-format=1.2]}}
\toprule
& \multicolumn{4}{c}{\textbf{Haiku 3 v1}} & \multicolumn{4}{c}{\textbf{Sonnet 3.5 v2}} & \multicolumn{4}{c}{\textbf{Sonnet 3.7 v1-standard}} & \multicolumn{4}{c}{\textbf{Sonnet 3.7 v1-think}} \\
\cmidrule(lr){2-5} \cmidrule(lr){6-9} \cmidrule(lr){10-13} \cmidrule(l){14-17}
\textbf{Experiment} & {\textbf{Prec}} & {\textbf{Rec}} & {\textbf{F1}} & {\textbf{Acc}} & {\textbf{Prec}} & {\textbf{Rec}} & {\textbf{F1}} & {\textbf{Acc}} & {\textbf{Prec}} & {\textbf{Rec}} & {\textbf{F1}} & {\textbf{Acc}} & {\textbf{Prec}} & {\textbf{Rec}} & {\textbf{F1}} & {\textbf{Acc}} \\
\midrule
\rowcolor{gray!10}
\makecell[l]{Simple Instruction \\ (Baseline)} & 0.76 & 0.42 & 0.33 & 0.42 & 0.78 & 0.19 & 0.29 & 0.19 & 0.47 & 0.94 & 0.62 & 0.60 & 0.67 & 0.88 & 0.76 & 0.78 \\
\makecell[l]{Human Optimized\\Instruction} & 0.59 & 0.54 & 0.39 & 0.42 & 0.66 & 0.42 & 0.35 & 0.42 & 0.67 & 0.77 & 0.71 & 0.80 & 0.78 & 0.88 & 0.85 & 0.85 \\
\rowcolor{gray!10}
\makecell[l]{LLM Optimized\\Instruction} & 0.78 & 0.49 & 0.44 & 0.49 & 0.73 & 0.74 & 0.73 & 0.74 & 0.86 & 0.86 & 0.86 & 0.86 & 0.86 & 0.89 & 0.87 & 0.87 \\
\bottomrule
\end{tabular}
\end{center}

\vspace{1em}

\FloatBarrier 
\section{Performance Comparison of Human and LLM Rationales Across Different Demonstration Sizes} \label{app:performance_rationales}
We compared the performance metrics (Precision, Recall, F1-score, and Accuracy) between human-generated and LLM-generated rationales across different demonstration sizes (3 to 300 examples) using four LLM models (Anthropic models—Claude 3.7 Sonnet v1, Claude 3.5 Sonnet v2, and Claude 3 Haiku v1). For Claude 3.7 Sonnet specifically, our analysis included both standard inference operation and enhanced thinking mode with 10,000 reasoning tokens. All metrics reported are weighted averages across classes, where weights are proportional to class frequencies in the dataset. Our results showed that LLM-generated rationales generally outperform human rationales and improve with more demonstrations, particularly for Claude Sonnet 3.7 v1.

\begin{center}
\captionof{table}{Performance Comparison of Human and LLM Rationales Using Claude Haiku 3}
\label{tab:haiku3_rationales_comparison}
\setlength{\tabcolsep}{4pt} 
\begin{tabular}{l*{8}{S[table-format=1.2]}}
\toprule
& \multicolumn{2}{c}{\textbf{Precision}} & \multicolumn{2}{c}{\textbf{Recall}} & \multicolumn{2}{c}{\textbf{F1-score}} & \multicolumn{2}{c}{\textbf{Accuracy}} \\
\cmidrule(lr){2-3} \cmidrule(lr){4-5} \cmidrule(lr){6-7} \cmidrule(l){8-9}
\textbf{Demonstration Count} & {\textbf{Human}} & {\textbf{LLM}} & {\textbf{Human}} & {\textbf{LLM}} & {\textbf{Human}} & {\textbf{LLM}} & {\textbf{Human}} & {\textbf{LLM}} \\
\midrule
\rowcolor{gray!10}
Few-shot (3) & 0.75 & 0.82 & 0.53 & 0.38 & 0.49 & 0.33 & 0.53 & 0.38 \\
Few-shot (10) & 0.74 & 0.81 & 0.54 & 0.55 & 0.52 & 0.55 & 0.54 & 0.55 \\
\rowcolor{gray!10}
Many-shot (50) & 0.60 & 0.80 & 0.59 & 0.70 & 0.59 & 0.72 & 0.59 & 0.70 \\
Many-shot (100) & 0.60 & 0.83 & 0.65 & 0.73 & 0.53 & 0.75 & 0.65 & 0.73 \\
\rowcolor{gray!10}
Many-shot (200) & {-} & 0.84 & {-} & 0.79 & {-} & 0.80 & {-} & 0.79 \\
Many-shot (300) & {-} & 0.86 & {-} & 0.82 & {-} & 0.83 & {-} & 0.82 \\
\bottomrule
\end{tabular}
\end{center}

\vspace{1em} 

\begin{center}
\captionof{table}{Performance Comparison of Human and LLM Rationales Using Claude Sonnet 3.5 v2}
\label{tab:sonnet35_rationales_comparison}
\setlength{\tabcolsep}{4pt} 
\begin{tabular}{l*{8}{S[table-format=1.2]}}
\toprule
& \multicolumn{2}{c}{\textbf{Precision}} & \multicolumn{2}{c}{\textbf{Recall}} & \multicolumn{2}{c}{\textbf{F1-score}} & \multicolumn{2}{c}{\textbf{Accuracy}} \\
\cmidrule(lr){2-3} \cmidrule(lr){4-5} \cmidrule(lr){6-7} \cmidrule(l){8-9}
\textbf{Demonstration Count} & {\textbf{Human}} & {\textbf{LLM}} & {\textbf{Human}} & {\textbf{LLM}} & {\textbf{Human}} & {\textbf{LLM}} & {\textbf{Human}} & {\textbf{LLM}} \\
\midrule
\rowcolor{gray!10}
Few-shot (3) & 0.74 & 0.88 & 0.69 & 0.79 & 0.70 & 0.80 & 0.69 & 0.79 \\
Few-shot (10) & 0.79 & 0.91 & 0.79 & 0.88 & 0.79 & 0.89 & 0.79 & 0.88 \\
\rowcolor{gray!10}
Many-shot (50) & 0.77 & 0.94 & 0.78 & 0.93 & 0.77 & 0.93 & 0.78 & 0.93 \\
Many-shot (100) & 0.79 & 0.89 & 0.76 & 0.86 & 0.76 & 0.87 & 0.76 & 0.86 \\
\rowcolor{gray!10}
Many-shot (200) & {-} & 0.92 & {-} & 0.92 & {-} & 0.92 & {-} & 0.92 \\
Many-shot (300) & {-} & 0.93 & {-} & 0.93 & {-} & 0.93 & {-} & 0.93 \\
\bottomrule
\end{tabular}
\end{center}

\vspace{1em} 

\clearpage
\begin{center}
\captionof{table}{Performance Comparison of Human and LLM Rationales Using Claude Sonnet 3.7 v1-standard}
\label{tab:sonnet37_standard_rationales_comparison}
\setlength{\tabcolsep}{4pt} 
\begin{tabular}{l*{8}{S[table-format=1.2]}}
\toprule
& \multicolumn{2}{c}{\textbf{Precision}} & \multicolumn{2}{c}{\textbf{Recall}} & \multicolumn{2}{c}{\textbf{F1-score}} & \multicolumn{2}{c}{\textbf{Accuracy}} \\
\cmidrule(lr){2-3} \cmidrule(lr){4-5} \cmidrule(lr){6-7} \cmidrule(l){8-9}
\textbf{Demonstration Count} & {\textbf{Human}} & {\textbf{LLM}} & {\textbf{Human}} & {\textbf{LLM}} & {\textbf{Human}} & {\textbf{LLM}} & {\textbf{Human}} & {\textbf{LLM}} \\
\midrule
\rowcolor{gray!10}
Few-shot (3) & 0.48 & 0.62 & 0.54 & 0.70 & 0.49 & 0.66 & 0.54 & 0.70 \\
Few-shot (10) & 0.76 & 0.83 & 0.62 & 0.83 & 0.68 & 0.83 & 0.62 & 0.83 \\
\rowcolor{gray!10}
Many-shot (50) & 0.77 & 0.90 & 0.66 & 0.85 & 0.71 & 0.85 & 0.66 & 0.85 \\
Many-shot (100) & 0.81 & 0.89 & 0.73 & 0.83 & 0.77 & 0.84 & 0.73 & 0.83 \\
\rowcolor{gray!10}
Many-shot (200) & {-} & 0.93 & {-} & 0.91 & {-} & 0.92 & {-} & 0.91 \\
Many-shot (300) & {-} & 0.96 & {-} & 0.95 & {-} & 0.95 & {-} & 0.95 \\
\bottomrule
\end{tabular}
\end{center}

\vspace{1em} 

\begin{center}
\captionof{table}{Performance Comparison of Human and LLM Rationales Using Claude Sonnet 3.7 v1-think}
\label{tab:sonnet37_thinking_rationales_comparison}
\setlength{\tabcolsep}{4pt} 
\begin{tabular}{l*{8}{S[table-format=1.2]}}
\toprule
& \multicolumn{2}{c}{\textbf{Precision}} & \multicolumn{2}{c}{\textbf{Recall}} & \multicolumn{2}{c}{\textbf{F1-score}} & \multicolumn{2}{c}{\textbf{Accuracy}} \\
\cmidrule(lr){2-3} \cmidrule(lr){4-5} \cmidrule(lr){6-7} \cmidrule(l){8-9}
\textbf{Demonstration Count} & {\textbf{Human}} & {\textbf{LLM}} & {\textbf{Human}} & {\textbf{LLM}} & {\textbf{Human}} & {\textbf{LLM}} & {\textbf{Human}} & {\textbf{LLM}} \\
\midrule
\rowcolor{gray!10}
Few-shot (3) & 0.72 & 0.81 & 0.65 & 0.81 & 0.62 & 0.79 & 0.65 & 0.80 \\
Few-shot (10) & 0.70 & 0.81 & 0.67 & 0.81 & 0.66 & 0.81 & 0.67 & 0.81 \\
\rowcolor{gray!10}
Many-shot (50) & 0.70 & 0.93 & 0.67 & 0.92 & 0.66 & 0.92 & 0.67 & 0.92 \\
Many-shot (100) & 0.74 & 0.96 & 0.74 & 0.96 & 0.74 & 0.96 & 0.74 & 0.96 \\
\rowcolor{gray!10}
Many-shot (200) & {-} & 0.97 & {-} & 0.96 & {-} & 0.97 & {-} & 0.97 \\
Many-shot (300) & {-} & {-} & {-} & {} & {-} & {} & {-} & {-} \\
\bottomrule
\end{tabular}
\end{center}

\end{document}